# Research Experience of an Undergraduate Student in Computer Vision and Robotics

Ayush V. Gowda, Dr. Juan D. Yepes, and Dr. Daniel Raviv

**Abstract**

This paper focuses on the educational journey of a computer engineering undergraduate student venturing into the domain of computer vision and robotics. It explores how optical flow and its applications can be used to detect moving objects when a camera undergoes translational motion, highlighting the challenges encountered and the strategies used to overcome them. Furthermore, the paper discusses not only the technical skills acquired by the student but also interpersonal skills as related to teamwork and diversity. In this paper, we detail the learning process, including the acquisition of technical and problem-solving skills, as well as out-of-the-box thinking.

The student joined this project with minimal robotics knowledge and only a basic understanding of computer vision. He learned about theoretical mathematical algorithms developed prior to his joining the project and was shown some existing Python and Excel simulations. After learning the theory, the student assembled a Hiwonder JetAuto Pro Jetson Nano robot, created an artificial 3D environment, developed a Python OpenCV program, implemented and verified the theory and simulations. He also recorded and processed relevant videos.

The student was part of a team consisting of a professor, an Electrical Engineering PhD student, and other undergraduate students. During their weekly meetings, they discussed various ideas for solving the problems they faced, employing both divergent and convergent thinking, and shared progress on their respective projects. These meetings facilitated the exchange of ideas, yielding multiple practical solutions. One such solution involved combining a mobile phone with the Hiwonder robot to capture and synchronize accelerometer and gyroscope data.

This paper traces the student's exploration, which was not limited to just classroom-based learning but also extended into hands-on experimentation using state-of-the-art robotic systems. We are exploring how to expand this success to many students interested in hands-on research.

The implementation of this research project culminated in a research paper submitted to a peer-reviewed conference.

1. **Introduction**

    Computer Science and Computer Engineering have emerged as defining forces of the 20th and 21st centuries, leaving their mark on human history with their levels of innovation and rapid progress [1]. These fields have not only revolutionized the way we interact with the world around us, but have also become attractive career prospects, supplying high-paying opportunities and intriguing projects [2].

    Among the various branches of these disciplines, Computer Vision has recently garnered significant attention due to its ability to mimic human-like perception using computing technology. By employing algorithms and processing data, it enables machines to comprehend and engage with the visual world. This has broadened the use of computers in fields that are typically reliant on human visual and processing skills such as transportation, manufacturing, and healthcare.

    Recognizing the immense potential for students that go into this field, this paper explores the educational trajectory of a Computer Engineering and Science student, in Computer Vision. It aims to highlight the potential for students in Computer Science and Engineering, focusing on the process of the student's progressive mastery of this field. Initially unfamiliar with the field, the student progressed to develop a low-complexity system for a robot to detect external motion unrelated to the robot's own motion while in transit, and publish a conference paper on the methodology.

    Moreover, this paper seeks to provide educators with insights to more effectively guide their students to success in Computer Science and Engineering, such as addressing the challenge of bridging the gap between aspirational goals, practical limitations of technology, and students' knowledge base. This mismatch is often one of the reasons for a "hump"- a phase where students often face discouragement and lose motivation. However, overcoming this "hump" is crucial for students to truly understand the concepts they are working with and learn how to deal with

similar situations in the future. This paper aims to give a detailed view into the learning process, triumphs, and pitfalls of an undergraduate student to allow educators to more effectively help their students.

## 2. Educational Goal

The goal of this project was to determine how an undergraduate student would approach and work on a high-level topic with minimal prior experience. Before joining the project, the student had some experience with Python equivalent to an introductory class, but was unfamiliar with visual looming. For this project, he learned visual looming as described in the paper by Raviv and Joarder (2000) [3].

Our aim was to observe the student's learning journey as the undergraduate student would grasp the principles of visual learning, develop a Python implementation of the principles, be given a problem along with a theoretical solution, craft a script demonstrating this solution, and finally integrate this script into a robot to prove its effectiveness. The decision to provide the undergraduate student with the theoretical solution to the problem was intentional and based on a shift in focus from previous studies. While earlier studies, such as Macri et al. (2023) [4] focused on helping students develop solutions to problems and bringing those solutions to reality, this paper aims to find and teach strategies to deal with the challenges and obstacles students will face in every other aspect of project development. This will show how various strategies can be implemented throughout development to proactively address challenges. This preemptive approach helps in maintaining student motivation and ensures success, by providing guidance and support tailored to the needs at each phase.

Overall there are four main educational goals in this paper:

Understanding Visual Looming - This aspect focuses on educating the student about the fundamental principles of visual looming. It involves comprehending and quantifying how objects appear to grow larger as they approach closer, and how to utilize this phenomenon.

Proficiency in Technical Tools - The student will learn how to utilize industry-standard technologies and tools to assist them during project development.

  - Programming with Python
  - Employing libraries like OpenCV and NumPy
  - Utilizing Git and GitHub
  - Operating Linux

Communication and Information Dissemination - Given the student will initially be inexperienced with the project, and in a new environment with new mentors, this element is vital. For example, weekly meetings with mentors, while useful for the student to clear up doubts, will not provide the student with much value if the student is unable to articulate where their problem lies.

Learning how to Learn -  The student will improve their learning and working methods week over week, allowing them to improve their ability to grasp and build upon information. This will help them become more productive and efficient as a researcher, overcoming any "humps" they come across and also develop strategies to help them overcome "humps" throughout their life.

In summary, the educational goals of this project were designed to go beyond teaching technical skills, and focus on enhancing the student's overall proficiency. The aim was to prepare them with a diverse skill set, fostering adaptability and readiness for professional environments beyond the classroom [5].

3. **Visual Looming**

   A. **Introduction to Team and Initial Learning Phase**
   Upon joining the team, the student was introduced to an environment free from immediate productivity pressures. This period was designated for learning, with minimal expectations for innovation or creation. This was made abundantly clear to the student to minimize any pressures on them. In their first meeting, the student received Dana H. Ballard's book "Computer Vision" [6] to help them acquaint themselves with the overarching topic at their own pace [7]. This approach emphasized understanding over immediate results, allowing the student to absorb the material, and seek clarification on confusing topics. The student was advised to not focus on memorizing the entire book. Instead, they were encouraged to explore the concepts at their own pace and they were encouraged to ask questions about topics that they found interesting without worrying about initial confusion on any topics. This approach aimed to establish a solid theoretical foundation, crucial for building upon it, and get the student interested in the subject-matter.

### B. Transition to Hands-On Learning and Focus on Visual Looming

Then the student's learning path shifted to a practical approach under a PhD student's mentorship, focusing specifically on the field of visual looming. This concept describes the quirk of objects moving toward a camera to appear to move towards the edges, and increase in size while doing so. He was taught the principles behind it and educated on using 2D and visual looming to calculate whether an object is moving towards or away from a point. Here the student was given their first task, implementing the Farneback model in a python script, a technique for visual looming tracking [8]. This task required them to record videos to test and experiment with the script, offering a practical approach to understanding and applying the concept.

### C. Mentor-Mentee Communication Dynamics

For any new relationship, both parties understanding how to effectively communicate with one another is vital as it facilitates smoother interactions. Initially, the student was hesitant to ask questions, however this was addressed by explaining the objective, and how we wanted to remove as many barriers towards that as possible. This transparency cultivated a trusting and open environment, enabling the student to gain confidence and become more proactive in asking questions.

Despite this progress, communication clarity became the area for improvement. The student had difficulty clearly articulating questions in a way the mentors could understand, leading to confusion and misunderstandings. To tackle this, mentors emphasized the importance of enhancing communication skills. They encouraged the student to prepare questions in advance, practice adding context, and be explicit in their communication. This strategy led to more effective mentoring sessions, enabling a focus on specific areas of confusion and reducing meeting fatigue. As a result, meetings became more efficient, with energy primarily directed towards learning and explaining material. This not only streamlined the sessions, but also helped the team to engage more effectively in problem-solving, ensuring a focus on areas needing the most attention and avoiding unnecessary repetition.

### D. Visual Looming Theory

Visual looming refers to the phenomenon where objects appear to increase in size as they approach closer to the observer.

To effectively analyze visual looming within a three-dimensional space, it is easier to employ a coordinate system that mirrors the functionality and perspective of a camera. For this polar coordinates, consisting of theta ($\theta$), phi ($\varphi$), and radius (r), were chosen.

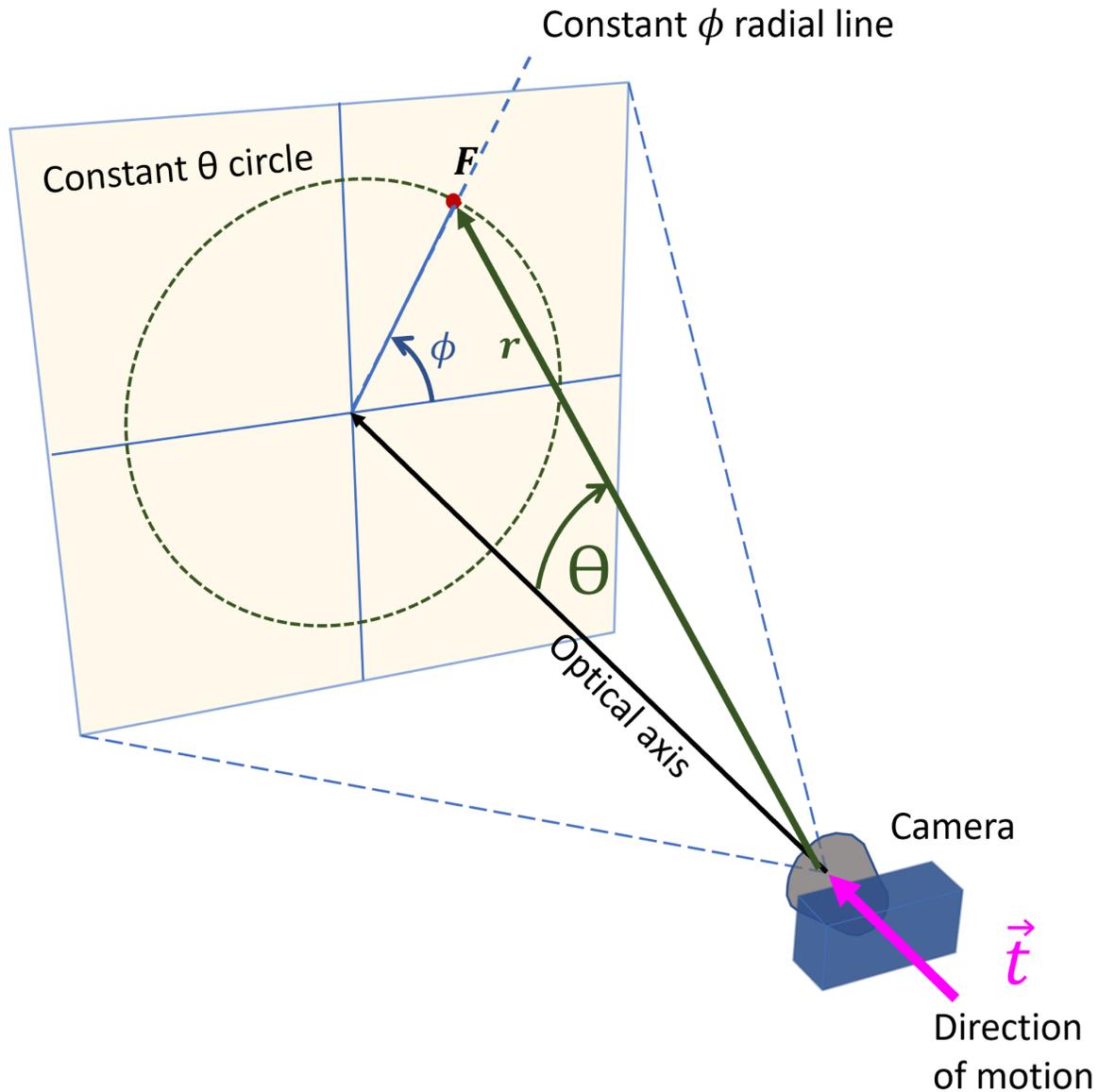

*Figure 1 [9]: Diagram showing coordinate systems*

1. Theta (θ): Represents the horizontal angle, theta is analogous to compass directions in a two-dimensional plane. It describes the left-to-right orientation of the camera's field of view.

2. Phi (φ): Represents the vertical angle, is perpendicular to theta. It signifies the up-and-down tilt of the camera, thereby accounting for the vertical aspect.

3. Radius (r): Denotes the distance from the camera to the point in space where the theta and phi angles intersect. Essentially measuring how far an object is from the observer or the camera lens.

Optical flow, another concept, allows us to measure visual looming, refers to the flow/movement of objects from one frame to another [10]. As it is inherently camera-based, the horizontal (x-axis) and vertical (y-axis) movement observed in pixels can be translated into corresponding theta and phi values. This translation allows for a more comprehensive and quantifiable understanding of object movement within the camera's field of view, particularly in analyzing the visual looming effect.

In scenarios involving linear or rectilinear motion, such as a vehicle moving directly towards the camera on a street, the relationship between theta and phi becomes increasingly direct. The degree of this change is directly proportional to the vehicle's velocity and its distance from the camera, embodying the concept of visual looming. By analyzing changes in theta and phi, one can infer the direction, speed, and thus the trajectory of an object's movement.

In summary, the use of polar coordinates with optical flow provides a low computation framework for quantifying visual looming and movement with a single 2D camera.

4. **Problem and Application**

A. **Problem**

After learning the basics of visual looming and creating a sample project, the student was presented with a new challenge. The task was to apply their understanding of visual looming and optical flow to develop a practical method for determining the movement of objects relative to an observer moving in straight, rectilinear motion. This involved using the ratio of two angles, phi ($\varphi$) and theta ($\theta$), to identify the dynamics of object movement.

B. **Application Theory**

The proposed application uses the Polar Coordinate system, theta ($\theta$), phi ($\varphi$), and radius (r). This problem focuses on an observer only moving in a straight line, and the principle of visual looming, where objects appear larger as they approach the observer, causing both $\theta$ and $\varphi$ to increase proportionally. By analyzing the ratio of $\theta$ to $\varphi$, the solution can differentiate between objects that follow, and those that deviate from expected looming behavior—where the growth in width and height should follow a specific pattern across the observer's field of view and between frames. This method allows for the clear identification of independently moving objects within the observer's environment. This is because the objects' size does not change at the same rate as the background, and the specific pattern that all stationary objects follow, making them stand out. This approach stands out because in previous models, both the object and background would appear to move and generate optical flow. This would make it a challenge to find one moving object, let alone multiple moving objects [11].

5. **The "Hump"**

   **A. Climbing The "Hump"**

The student began by programming a script that could perform the proposed solution, but soon encountered their first significant obstacle, and got stuck on this "hump". This challenge arose from a lack of Python skills and specific knowledge about the Farneback model. The student had a hard time collecting the phi ($\varphi$) and theta ($\theta$) values, and the student's lack of experience led to them feeling overwhelmed and unable to make progress.

Addressing procrastination by understanding and addressing its underlying reasons significantly boosts productivity and enthusiasm. This skill is crucial, not only for educators in guiding students, but also for students to develop independently. Self-awareness and proactive habits in overcoming procrastination can greatly benefit students throughout their lives, allowing them to work through any problem that they come across.

In this situation, the student expressed feeling stuck, and the mentors drew up a plan with the student on how to get past this roadblock. The plan involved initially concentrating on improving the student's Python skills, particularly through projects that would teach the student how to effectively manipulate matrices using NumPy, given that the output of the Farneback model is in matrix form. Once the student gained familiarity with NumPy, the mentors had the student practice extracting matrices from the Farneback model. From here, the mentors had the student combine his new knowledge on manipulation and extraction of matrices to get the student up and over the hump that he was stuck on.

Through this approach, the mentors successfully broke the student out of the procrastination cycle. The mentors helped the student break the project down into smaller, more manageable segments, and then link these segments together as the student grew more confident in their abilities. By identifying patterns that led to procrastination, the student not only overcame the immediate challenge, but also learned to segment and tackle problems in other areas of the project independently. This strategy prevented the onset of procrastination, significantly boosting the student's enthusiasm, and contributing to the project's overall success.

**B. Going Around the "Hump"**

Having completed the programming aspect, the student now had to configure the robot to move in specific patterns while simultaneously capturing video. This aimed to demonstrate the applicability of the Visual Looming Transform in the real world. However, the student encountered difficulties saving recordings using the HiWonder JetAuto Pro robot's camera module. These issues were largely due to the student's inexperience with Linux distributions and the HiWonder camera modules. Although the student could use the segmented method to break down this challenge, the student instead adapted to the situation and screen recorded the camera module's display, saving that video instead.

This experience highlights the importance for students to have a clear end goal, the ability to adapt, and an understanding of how each step contributes to achieving the final objective. This segment of the project was broken down into these steps: capturing video input, running the simulation, and outputting results. When direct saving attempts failed to capture the video input, the student adapted and instead began screen recording the display, showing the camera's view instead. But this was only possible because the student understood why they were doing what they were doing. This creative solution not only saved hours that would've been spent sifting through code, but also kept the student motivated, maintaining momentum on the project.

## 6. Implementation

### A. Hardware:

- *Tri-Fold Board:* Used as the background of the videos. Covered in random scribbles and dots made with colored pencils and crayons, enhancing the recorded video's visual complexity.

- *JetAutoPro Robot by HiWonder Shenzhen:* The primary hardware platform used for testing, featuring an arm-mounted camera.

- *Robot Arm Camera:* Configured to move parallel to the floor, and perpendicular to the Tri-Fold board.

- *PVC Pipes and Two-Way Adaptors:* Used to assemble a dedicated track, measuring 5 feet in length, and 11 inches in width, with four two-way adaptors.

- *USB Dell Keyboard:* Used for external control of the robot, ensuring maneuverability.

- *Logitech Bluetooth Mouse:* Employed alongside the keyboard for external control of the robot.

- *iPhone:* Used in conjunction with PyPhox to capture ground-truth movement data.

- *Laptop:* Used to run simulations, and run the Visual Looming Transform script

### B. Software:

- *Built-in Modules on the JetAutoPro Robot:* Original modules adapted for precise control, responding to keyboard inputs for movement commands. Displaying camera feed.

- *SimpleScreenRecorder:* Utilized on the robot to capture the camera feed.

- *PhyPhox Application on iPhone:* Awarded application known for its accelerometer functionality, and used to gather comprehensive motion data.

- *Visual Looming Transform Script:* Created based on the theory described in the above sections and [9].

- *PyCharm IDE:* Integrated Development Environment built for Python by JetBrains

**C. Data Collection**

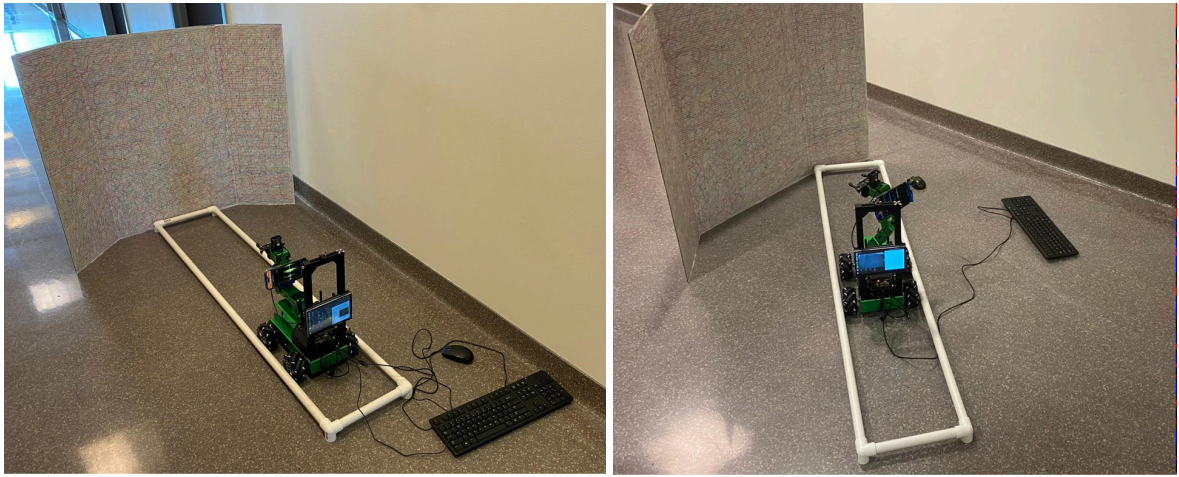

*Figure 2&3: These two images depict the robot's movement aligned on the track and the Tri-Fold board as the background. In both images the camera is parallel to the ground and perpendicular to the board. On the left (Fig. 2), the robot is moving perpendicular to the Tri-Fold board. On the right (Fig. 3), the robot is moving slanted to the Tri-Fold board.*

Real-world testing consisted of two distinct movement patterns. First, the HiWonder Shenzhen's JetAutoPro robot moving in a straight or nearly straight line with the arm mounted camera oriented forward, and then the robot advancing in a direction that was slanted relative to the camera's orientation. Both types of experiments were performed using a PVC track to ensure the robot was moving as intended. The robot features a camera mounted on its arm, aligned parallel to the ground, and perpendicular to a static-patterned Tri-Fold board as seen in Figures 2 and 3 [9]. The speed and video capture were regulated via adapted HiWonder modules. Calibration between the recorded video and the accelerometer was done using Z-axis data which was collected by the researcher physically lifting the robot up and down the Z-axis. This did not affect the experimental results because Z axis data was not used to determine experimental results.

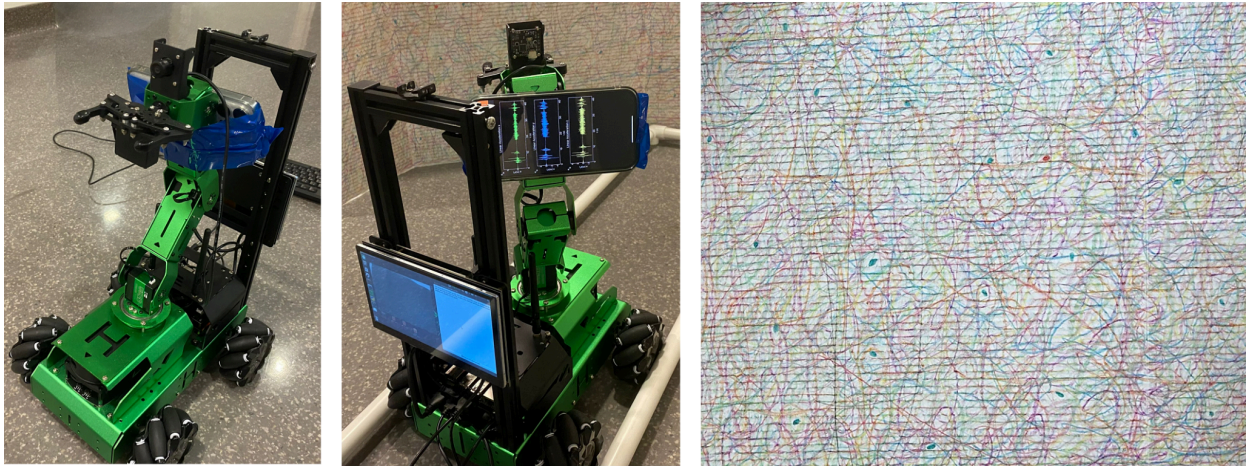

*Figure 4 [9]: The JetAutoPro robot is shown with the arm mounted camera on the left. In the middle is the back view of the robot, and the iPhone displaying PhyPhox. The Tri-Fold board with the simulated static is shown on the right.*

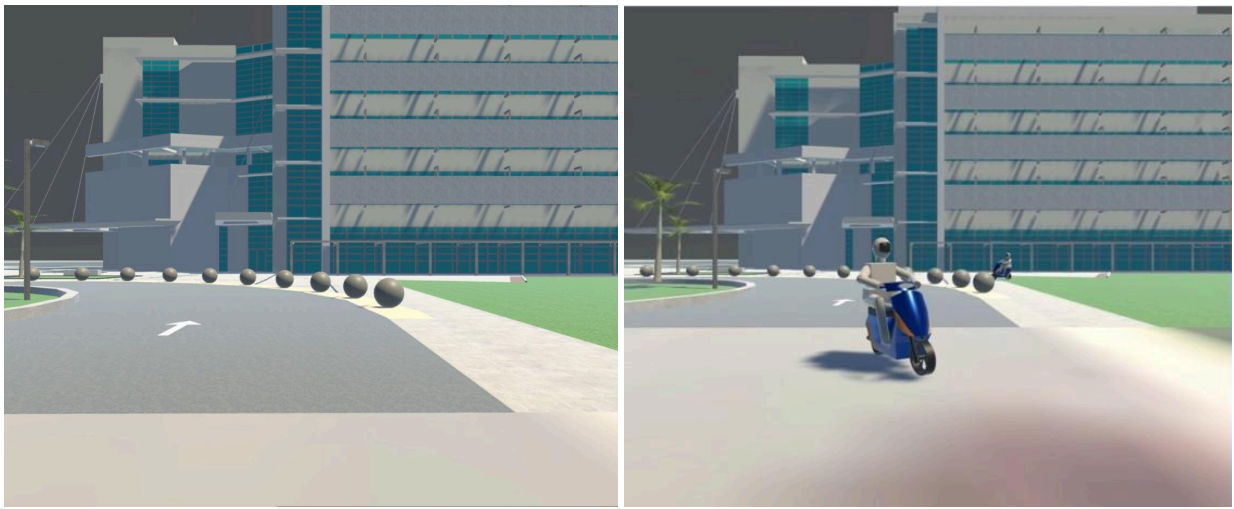

*Figure 5 & 6: Both images are stills from a unity simulation with a stationary background, and the observer moving forward. Left (Fig. 5): Only background. Right (Fig. 6) Background with two motorists approaching the observer.*

Simulation testing was performed on Unity [12], and consisted of two types of tests in which the observer moved straight ahead. In one (Fig. 6) [9], there would be vehicles moving relative to the observer, and in the other (Fig. 5) [9], the environment was stationary without any vehicles. Simulation testing was not done by the student independently, but it is included to show the student's Visual Looming Transform script working as intended in a perfect simulated environment. This is because the research is ongoing and real-world results have yet to be published.

### D. Data Analysis

The video recordings were preprocessed to ensure uniform dimensions by cropping. For processing, simulation videos utilized the RAFT [13] method, whereas real-world testing videos employed the Farneback method. On the HiWonder JetAuto Pro, the onboard NVIDIA Jetson Nano struggled with the demands of RAFT due to its high computational requirements. RAFT, being a more recent and accurate software, is also more computationally intensive. In contrast, Farneback's lesser computational load enabled concurrent program execution, robot movement, camera input acquisition, and result viewing.

Both RAFT and Farneback are techniques for identifying, visualizing, and quantifying optical flow. After collecting the quantified data from RAFT and Farneback, the student then used the Visual Looming Transform script, which divides the values of phi and theta from every pixel in each frame to get the specific pattern that is always created by the background while the observer moves. This allows the researcher to see if any objects deviate from that pattern, clearly showing that those objects are moving relative to the stationary background.

The computed values were then compared with the ground-truth accelerometer data obtained from both the iPhone and the simulator. This comparative analysis aimed to gauge the accuracy of using optical flow for motion detection and analysis. However, this segment of the research is still ongoing, and results will not be present in this paper.

### E. Results and Discussion

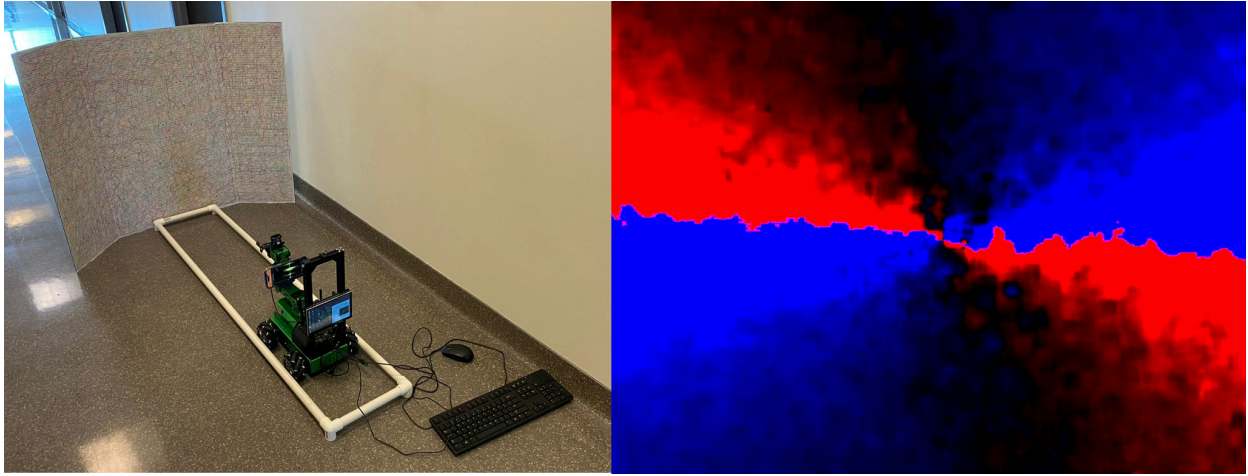

*Figure 7: On the left, the robot is moving perpendicular to the Tri-Fold board. On the right, the Visual Looming transformation from this motion is shown.*

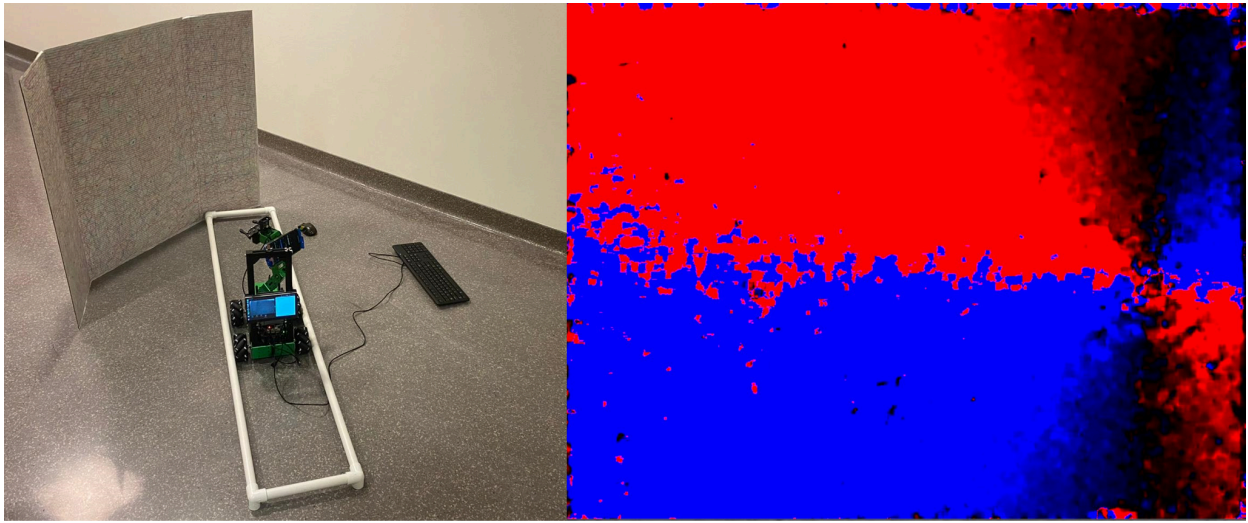

*Figure 8: On the left, the robot is moving slanted to the Tri-Fold board. On the right, the Visual Looming transformation from this motion is shown.*

Real-world tests demonstrated that the transformation results closely resembled the theoretical pattern, highlighting this method's practical capability despite the noise of real-world conditions. In Figure 7, this method's output can be seen, with the image made up of four quadrants around the central point which is the Focus of Expansion (FoE)[14]. The quirks of this image can be explained by division, which causes the middle vertical area to appear darker. This effect occurs because the calculation involves dividing a smaller number by a larger number, which approaches zero, leading to a darker and less intense color representation. In contrast, the central horizontal segment has a brighter appearance, due to dividing a larger number by a smaller one.

The pattern remains consistent even when the robot moves diagonally, as shown in Figure 8 [9]. Despite the robot's slanted trajectory, maintaining the camera perpendicular to the Tri-Fold surface ensures the clear separation of quadrants and shows the slanted advance by shifting the focus of expansion. This consistency across different images validates the Visual Looming transform's reliability as a universal guide. This allows us to determine if there are any moving objects in a stationary background relative to an observer moving.

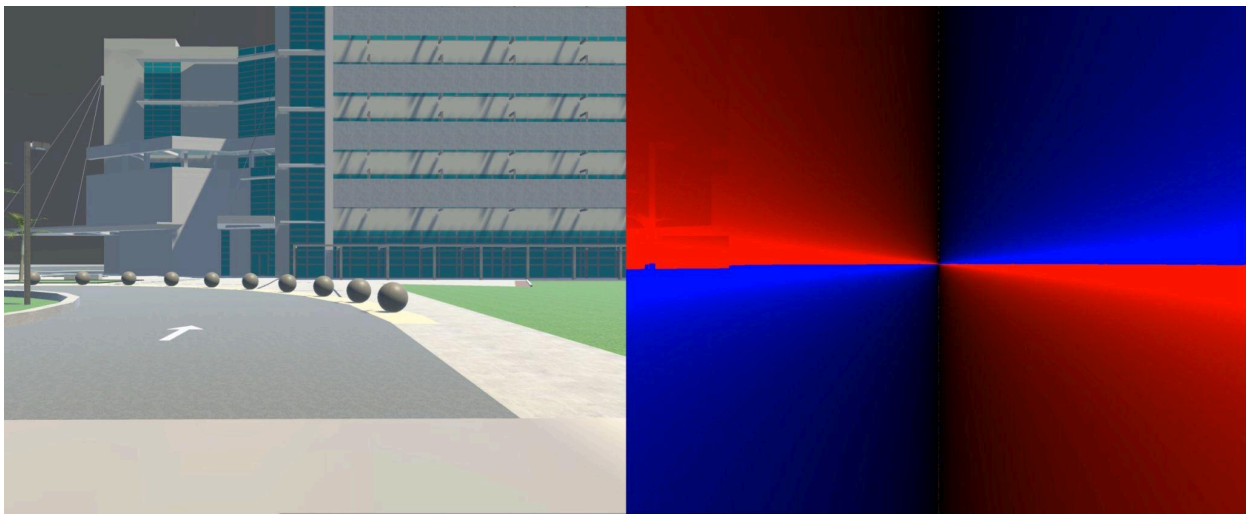

*Figure 9: On the left, the observer is moving forward toward a stationary background. On the right, the Visual Looming transformation from this motion is shown.*

Simulation testing further shows this concept. In Figure 9 [9], where the scene is devoid of moving objects it showcases only stationary background items, and the dark pillar in the Visual Looming transformation previously observed in real-world tests. Conversely, Figure 10 [9] introduces moving objects, which are immediately distinguishable due to their contrasting colors against the background. The efficiency and low implementation cost of this transform, needing just a single 2D camera and minimal computing resources, offer substantial potential for threat detection and obstacle avoidance.

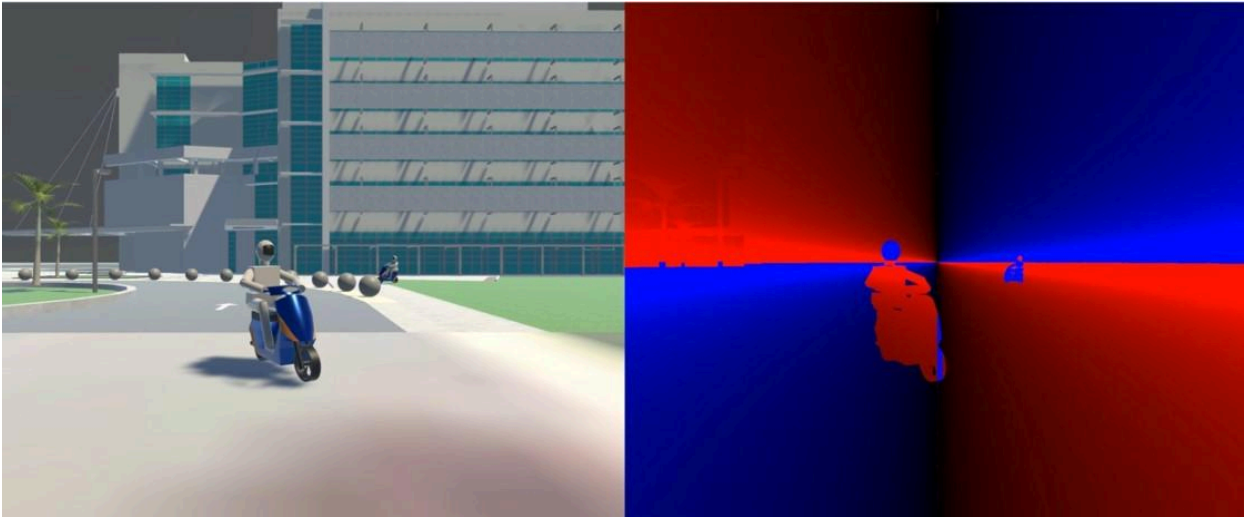

*Figure 10: On the left, the observer is moving forward toward a stationary background with two motorists. On the right, the Visual Looming transformation from this motion is shown.*

7. **Conclusion**

In conclusion, this paper has explored the educational journey of an undergraduate student in the field of Computer Science and Engineering. The student started out as a beginner having never heard of Visual Looming or Optical Flow, and only had a basic understanding of Python. However, by the end, the student had progressed to a competent individual that had built a full system from theoretical concepts to a moving robot running a script developed by the student.

Moreover, the student developed the ability to navigate through difficulties, effectively communicate, apply theoretical knowledge to practical problems, and continuously improve their skills highlighting the importance of fostering an encouraging attitude in educational settings. The focus on "learning how to learn" and improving oneself week over week is a critical takeaway from this experience, and provides the student with a lifelong strategy to tackle problems, which is invaluable in the ever-evolving field of technology.

This paper also emphasizes the role of mentors and educators in guiding students through their educational journey. By providing support, encouragement, and real-world problems to solve, educators can significantly enhance the learning experience, helping students to bridge the gap between theoretical knowledge taught in classrooms and practical applications done in the workspace. This approach not only prepares students for successful careers in Computer Science and Engineering, but also instills in them the confidence and skills needed to tackle future challenges.

As the authors, we hope this highlights the need for educational programs that are not just focused on imparting technical knowledge, but are also committed to developing the whole individual—equipping students with the skills, mindset, and adaptability required to thrive in the real world. Through such programs, students can emerge as competent professionals, innovative thinkers, and problem solvers who can contribute more to the world while also succeeding in their future endeavors.